\documentclass[letterpaper, 10 pt, conference]{ieeeconf}  

\IEEEoverridecommandlockouts                              

\usepackage{graphics} 
\usepackage{graphicx}
\usepackage{color}
\usepackage{amsmath} 
\usepackage{amssymb}  
\usepackage{makecell}
\usepackage{url}

\title{\LARGE \bf
Robust Model-Aided Inertial Localization for Autonomous Underwater Vehicles
}

\author{Sascha Arnold and Lashika Medagoda
\thanks{Sascha Arnold and Lashika Medagoda are with the German Research Center for Artificial Intelligence, Bremen, Germany
        {\tt\small \{sarnold,lmedagoda\}@dfki.de}}%
}

\begin{document}

\maketitle
\thispagestyle{empty}
\pagestyle{empty}

\linespread{0.96}

\begin{abstract}

This paper presents a manifold based Unscented Kalman Filter that applies a novel strategy for inertial, model-aiding and Acoustic Doppler Current Profiler (ADCP) measurement incorporation. The filter is capable of observing and utilizing the Earth rotation for heading estimation with a tactical grade IMU, and utilizes information from the vehicle model during DVL drop outs. The drag and thrust model-aiding accounts for the correlated nature of vehicle model parameter error by applying them as states in the filter. ADCP-aiding provides further information for the model-aiding in the case of DVL bottom-lock loss. Additionally this work was implemented using the MTK and ROCK framework in C++, and is capable of running in real-time on computing available on the \textit{FlatFish} AUV.
The IMU biases are estimated in a fully coupled approach in the navigation filter. Heading convergence is shown on a real-world data set. Further experiments show that the filter is capable of consistent positioning, and data denial validates the method for DVL dropouts due to very low or high altitude scenarios.

\end{abstract}

%
%
%
%


\section{Introduction}

Autonomous Underwater Vehicles (AUVs) have found applications in a variety of underwater exploration and monitoring tasks including high-resolution, geo-referenced optical/acoustic ocean floor mapping and measurements of water column properties such as currents, temperature and salinity \cite{yoerger1991autonomous}. An advantage of AUVs over other methods of ocean observation is the autonomy and decoupling from a surface vessel that a self-contained robot provides.

The ability to geo-reference, or to compute the absolute position in a global reference frame, is essential for AUVs for the purposes of path planning for mission requirements, registration with independently navigated information, or revisiting a previous mission. Geo-referenced navigation is often achieved by initializing the navigation solution to the GPS while on the surface and, once submerged, relaying on velocity measurements from a  Doppler Velocity Log (DVL).  When the water depth is less than the range of the DVL (a 300kHz DVL has a range of $\sim$200m), the DVL has continuous bottom lock throughout the mission.  The DVL sensor provides measurements of the seafloor relative velocity of the AUV.  By combining this information with an appropriate heading reference, the observations are placed in the global reference frame and integrated to facilitate underwater dead reckoning. The result is accuracies of 22m per hour (2\(\sigma\)) in position error growth attainable during diving and 8m per hour error growth (2\(\sigma\)) is possible if coupled with a navigation-grade (\(>\)\$100K) Inertial Measurement Unit (IMU) \cite{napolitano2004phins}.

In cases where the seafloor depth is greater than the DVL bottom lock range, transitioning from the surface to the seafloor presents a localization problem \cite{kinseynonlinear2014}, since both GPS and DVL are unavailable in the mid-water column. Traditional solutions include range-limited Long Base Line (LBL) acoustic networks requiring deployment, Ultra Short Base Line (USBL) navigation requiring a dedicated ship, or single range navigation from an acoustic beacon attached to a ship \cite{webster2012advances} or an autonomous surface vehicle (ASV) \cite{kinsey2013-auv-asv}. In addition to requiring dedicated infrastructure, acoustic positioning also suffers from multipath returns and the need to accurately measure the sound speed profile through the water column. Acoustic methods typically give \(\mathcal{O}({10m})\) accuracy at 1km ranges \cite{kinsey2006survey} \cite{mandt2001integrating}.

\begin{figure}[!t]
      \centering
      \includegraphics [width=0.45\textwidth] {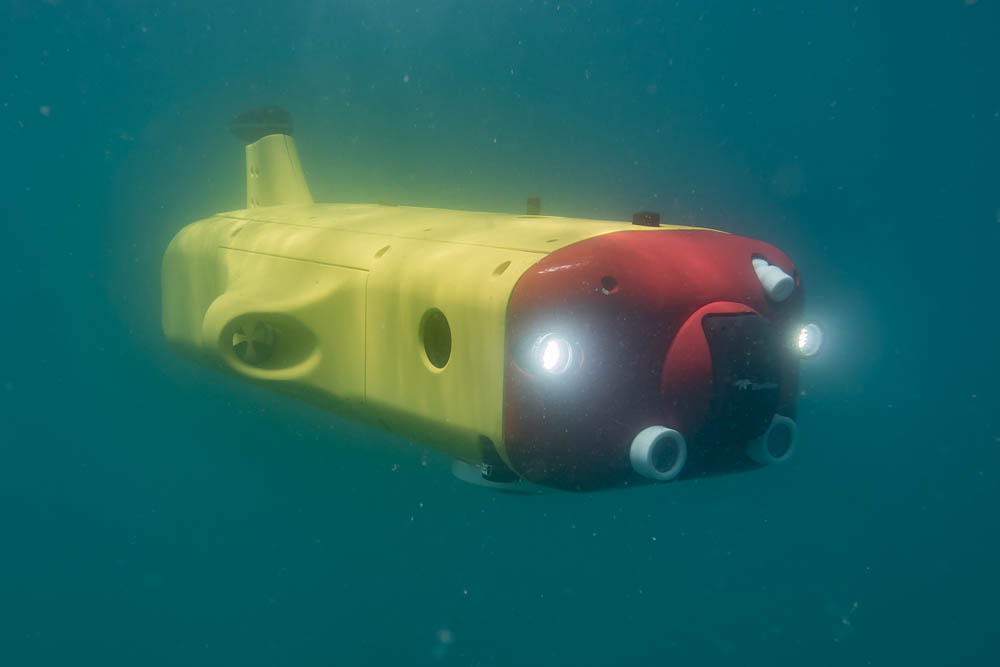}
      \caption{The \textit{FlatFish} AUV \cite{albiez2015flatfish} during sea trails. Image: Jan Albiez, SENAI CIMATEC}
      \label{fig:flatfish}
\end{figure}

IMUs provide a strap down navigation capability through providing body accelerations and rotation rates without external aiding such as GPS, acoustic ranging, or DVL velocities. However, IMUs quickly accumulate position errors, with an unaided tactical grade IMU (\(>\)\$10K) drifting at $\sim$100km per hour, and a navigation grade IMU drifting at $\sim$1km per hour \cite{titterton2004strapdown}. There also exists cases where DVL bottom-lock is not possible, when the altitude is very low, such as in inspection or docking scenarios.

In \cite{hegrenaes2011model}, a model-aiding Inertial Navigation System (INS) is applied with water-track from the DVL. Comparatively, the novel contributions of the work presented in this paper are as follows: 
\begin{enumerate}
	\item Utilizing and validating through experiment a manifold based Unscented Kalman Filter (UKF) which can observe and utilize the Earth rotation for heading estimation,
	\item Incorporating and validating a novel drag and thrust model-based aiding, which accounts for the systematic uncertainty in vehicle parameters by incorporating them as states in the UKF and
	\item Incorporating and validating the use of ADCP measurements in a novel form to further aid the estimation in cases of DVL bottom-lock loss.
\end{enumerate}

IMUs with low gyro bias uncertainty allow gyrocompassing by measuring the Earth rotation to estimate heading. The price range for navigation grade IMUs (as used in \cite{hegrenaes2011model}) with a low bias uncertainty are typically in the \(>\)\$100K USD price range. In this paper, the KVH1750 IMU, in the \(>\)\$10K USD price range, is utilized. In order for this price range IMU to be utilized, the biases are estimated in a fully coupled approach in the navigation filter. Real-world experiments with the \textit{FlatFish} AUV (Fig. \ref{fig:flatfish}) show that less than 1$^{\circ}$ (2$\sigma$) heading uncertainty is possible in the filter following an initialization within 15$^{\circ}$ of the true heading (possible from a magnetic sensor). Further experiments also show that the filter is capable of consistent positioning, and data denial validates the method for DVL dropouts due to very low or high altitude scenarios. Additionally this work was implemented using the MTK \cite{hertzberg2013integrating} and ROCK \cite{rock} framework in C++\footnote{The implementation is under open source license available on \url{https://github.com/rock-slam/slam-uwv_kalman_filters}}, and is capable of running in real-time on computing available on the \textit{FlatFish} AUV.

The work in this paper utilizes vehicle model-based aiding and the ADCP sensor for further ocean water current and vehicle velocity constraints. Model-aiding allows physics based constraints on the positioning, and the uncertainty in each parameter can be set to account for the systematic error associated with a system identification. Thus even a low accuracy system identification can still be used with this filter without resulting in filter overconfidence. Additionally, by modeling the vehicle parameters as time-varying, the model itself has become uncertain, as any small deviations in dynamics from the modeling equations can be absorbed by the time-varying parameters. ADCP-aiding in cases where DVL dropout would occur, due to being higher altitude than the bottom-lock range, also can aid the model by providing independent vehicle velocity constraints. ADCP also gives information regarding the surrounding water currents when there is a DVL dropout and the vehicle state estimation relies more on model-aiding.
Generally, inertial navigation is achieved using error-state filtering \cite{hegrenaes2011model}, but this is not necessary as is shown in this paper. This paper gives a more conceptually simplified approach, while also utilizing manifold methods \cite{forster2015manifold} to represent attitude, which is more general than other methods.


%

\section{Model-aided Inertial filter design}


%
%
%
%
%


Our filter design is conceptually simple, since we model all modalities in one filter and model the attitude of the vehicle as a manifold.
We utilize an Unscented Kalman Filter (UKF) since it doesn't require the Jacobians of the process or measurement models and can handle non-linearities better than an Extended Kalman Filter \cite{wan2000unscented}.

The attitude of the vehicle is an element of $SO(3)$, the group of orientations in $\mathbb{R}^3$. To directly estimate the attitude in the filter it can be either modeled by a minimal parametrization (Euler angles) or by a over-parametrization (quaternion or rotation matrix). A minimal parametrization has singularities, i.e. small changes in the state space might require large changes in the parameters. An over-parametrization has redundant parameters and needs to be re-normalized as required. In both cases it requires special treatment in the filter, which leads to a conceptually more complex filter design.
Representing the attitude as a manifold is a more general solution in which the filter operates on a locally mapped neighborhood of $SO(3)$ in $\mathbb{R}^3$ \cite{hertzberg2013integrating}.

\begin{table}[h]
\centering
\caption{Filter state}
\begin{tabular}{l|l}
\hline
\thead{Elements of \\ the state vector} & Description \\
\hline
$\mathbf{p}^n \in \mathbb{R}^3$		& Position of the IMU in the navigation frame \\
$\boldsymbol{\phi}^n \in \mathbb{R}^3$ 	& Attitude of the IMU in the navigation frame \\
$\mathbf{v}^n \in \mathbb{R}^3$ 	& Velocity of the IMU in the navigation frame \\
$\mathbf{a}^n \in \mathbb{R}^3$ 	& Acceleration of the IMU in the navigation frame \\
$\mathbf{M_{\text{sub}}} \in \mathbb{R}^{2\times3}$  & Inertia sub-matrix of the motion model \\
$\mathbf{D}_{l,\text{sub}} \in \mathbb{R}^{2\times3}$  & Linear damping sub-matrix of the motion model \\
$\mathbf{D}_{q,\text{sub}} \in \mathbb{R}^{2\times3}$  & Quadratic damping sub-matrix of the motion model \\
$\mathbf{v}_{c,v}^{n} \in \mathbb{R}^{2}$	& \begin{tabular}{@{}l@{}}Water current velocity surrounding \\ the vehicle in navigation frame\end{tabular} \\
$\mathbf{v}_{c,b}^{n} \in \mathbb{R}^{2}$	& \begin{tabular}{@{}l@{}}Water current velocity below the \\ vehicle in navigation frame\end{tabular} \\
$g^n \in \mathbb{R}$		& Gravity in the navigation frame \\
$\mathbf{b}_{\omega} \in \mathbb{R}^3$ 	& Gyroscope bias \\
$\mathbf{b}_{a} \in \mathbb{R}^3$ 	& Accelerometer bias \\
$\mathbf{b}_{c} \in \mathbb{R}^{2}$ 	& Bias in the ADCP measurements \\
\hline
\end{tabular}
\label{state_table}
\end{table}

Table \ref{state_table} shows the state vector of the filter as element of $\mathbb{R}^{43}$ and  gives a detailed description of the higher dimensional elements of the state vector. The navigation frame is North-East-Down (NED).
The body and IMU frames are x-axis pointing forward, y-axis pointing left and z-axis pointing up. 
In the filter design we consider the IMU frame not to be rotated with respect to the body frame.


\subsection{Inertial prediction equations}

The following equations describe the prediction models for position, velocity, acceleration and attitude, applying a constant acceleration model for translation and a constant angular velocity model for rotation:
\begin{equation}
\mathbf{p}^n_{t} = \mathbf{p}^n_{t-1} + \mathbf{v}_{t-1}^n \delta t
\label{eq:pred1}
\end{equation}
\begin{equation}
\label{vn}
\mathbf{v}^n_{t} = \mathbf{v}^n_{t-1} + \mathbf{a}_{t-1}^n \delta t
\end{equation}
\begin{equation}
\label{vn}
\mathbf{a}^n_{t} = \mathbf{a}^n_{t-1}
\end{equation}
\begin{equation}
\boldsymbol{\phi}^n_{t} = \boldsymbol{\phi}^n_{t-1} \boxplus [C^n_{b,t-1}(\boldsymbol{\omega}_{t-1}^{b} - \mathbf{b}_{\omega,t-1}) - \boldsymbol{\Omega}_{e}^{n} \delta t]
\label{eq:pred2}
\end{equation}

where $\mathbf{p}^n_{t}$ is the position of the IMU in the navigation frame at time $t$, $\mathbf{v}^n_{t}$ is the velocity of the IMU in the navigation frame, $\mathbf{a}^n_{t}$ is the acceleration of the IMU in the navigation frame, $\mathbf{C}^n_{b,t}$ is the coordinate transformation from body to navigation frame, $\boldsymbol{\phi}^n_{t}$ is the attitude of the IMU in the navigation frame, $\boldsymbol{\omega}^b_t$ is the rotation rates in the body frame, $\mathbf{b}_{\omega,t}$ is the gyroscope bias and $\boldsymbol{\Omega}_{e}^{n}$ is the Earth rotation in the navigation frame. The $\boxplus$ operator in \eqref{eq:pred2} is a manifold based addition, as defined in \cite{hertzberg2013integrating}.
Equations \eqref{eq:pred1} to \eqref{eq:pred2} each have corresponding prediction noise added.

The accelerometer measurements are handled with an update equation on the acceleration state as follows:

\begin{align}
\textbf{z}_{a}(t) = \mathbf{f}^b_t+ \mathbf{b}_{a,t} + \mathbf{C}^{n}_{b,t}\mathbf{g}^{n}_{t} + \nu_{a}
\end{align}

where $\mathbf{f}^b_t$ is the specific force acting on the vehicle at time $t$, $\mathbf{b}_{a,t}$ is the accelerometer bias and $\mathbf{g}^{n}_{t}$ is the gravity vector $\begin{bmatrix} 0, 0, g_t^n \end{bmatrix}^T$ in the navigation frame.
The gravity state is modeled applying a constant gravity model in order to refine the theoretical gravity according to the WGS-84 ellipsoid earth model starting with a small initial uncertainty.
The acceleration state in the filter allows both the accelerometer and model-aiding to act on the filter in a consistent fashion, without resorting to virtual correlation terms when an acceleration state does not exist, such as in \cite{hegrenaes2011model}.
Accelerometer and gyro bias terms are modeled as a first order Markov process as follows:

\begin{equation}
\dot{b} = -\frac{1}{\tau_{b}}(b-b_{0}) + \nu_{b}
\label{bias_equation}
\end{equation}
where $\tau_{b}$ is the expected rate change of the bias, $b_{0}$ is the mean bias value, and $\nu_{b}$ is a zero-mean normally distributed random variable with
\begin{equation}
\sigma_{b} = \sqrt{\frac{2f\sigma_{b\,drift}^2}{\tau_{b}}}
\end{equation}
where $\sigma_{b\,drift}$ is a bound to the bias drift and $f$ is the measurement frequency. The accelerometer and gyro bias are assumed to be zero mean.

\subsection{Model-aiding update equations}

In this section we show a model-aiding measurement function using a simplified vehicle motion model for which a subset of the parameter space is part of the filter state.
This allows the filter to refine the parameters at runtime and to account for the systematic uncertainty in the vehicle parameters.

The nonlinear equations for motion \cite{fossen2002marine} of a rigid body with 6 DOF can be written as
\begin{equation}
\label{eq:motion_fossen}
\boldsymbol{\tau} = \mathbf{M} \dot{\boldsymbol{\nu}} + \mathcal{C}(\boldsymbol{\nu}) \boldsymbol{\nu} + \mathbf{D}(\boldsymbol{\nu}) \boldsymbol{\nu} + \mathbf{g}(\mathbf{R}_{b}^n)
\end{equation}
where $\boldsymbol{\tau}$ is the vector of forces and torques, $\boldsymbol{\nu}$ is the vector of linear and angular velocities, $\mathbf{M}$ is the inertia matrix including added mass, $\mathcal{C}(\boldsymbol{\nu})$ is the Coriolis and centripetal matrix, $\mathbf{D}(\boldsymbol{\nu})$ is the hydrodynamic damping matrix and $\mathbf{g}(\mathbf{R}_{b}^n)$ is the vector of gravitational forces and moments given the rotation matrix from the body to the navigation frame $\mathbf{R}_{b}^n$.

\begin{equation}
\mathbf{g}(\mathbf{R}) = \begin{bmatrix} \mathbf{R}^{-1} \hat{k} (W-B) \\ \mathbf{r}_{G} \times \mathbf{R}^{-1}\hat{k}W - \mathbf{r}_{B} \times \mathbf{R}^{-1}\hat{k}B \end{bmatrix}
\label{eq:gravity}
\end{equation}
Equation~\eqref{eq:gravity} shows how the gravitational forces and moments are calculated given the weight $W$, buoyancy $B$, center of gravity $\mathbf{r}_{G}$ and center of buoyancy $\mathbf{r}_{B}$ of the vehicle,
where $\hat{k}$ is the unit vector $\begin{bmatrix} 0, 0, 1 \end{bmatrix}^T$.

We assume the Coriolis and centripetal forces as well as damping terms higher than second order are negligible for vehicles operating within lower speeds (typically below $1.5$ m/s). 
This allows us the define the measurement function for the forces and torques in the body frame from \eqref{eq:motion_fossen} as
\begin{equation}
\textbf{z}_{\boldsymbol{\tau}}(t) = \mathbf{M}_{t} \begin{bmatrix}\mathbf{a}_{t}^{b} \\ \boldsymbol{\alpha}_{t}^{b}\end{bmatrix} + \mathbf{D}(\begin{bmatrix}\mathbf{v}_{t}^{b} \\ \boldsymbol{\omega}^b_t\end{bmatrix},t) + \mathbf{g}(\mathbf{R}_{b,t}^n) + \nu_{\boldsymbol{\tau}}
\label{eq:meas_tau}
\end{equation}
where $\mathbf{a}_{t}^{b}$ is the linear acceleration, $\boldsymbol{\alpha}_{t}^{b}$ is the angular acceleration, $\mathbf{v}_{t}^{b}$ is the linear velocity and $\boldsymbol{\omega}^b_t$ is the angular velocity, all expressed in the body-fixed frame at time $t$. $\nu_{\boldsymbol{\tau}}$ is the random noise of the force and torque measurement, with a standard deviation given by the thruster manufacturer.

$\mathbf{a}_{t}^{b}$ can be computed given the acceleration in the navigation frame $\mathbf{a}_{t}^{n}$ as
\begin{equation}
\mathbf{a}_{t}^{b} = \mathbf{C}_{n,t}^{b}\mathbf{a}_{t}^{n} - \boldsymbol{\omega}^b_t \times (\boldsymbol{\omega}^b_t \times \mathbf{p}^b)
\label{eq:acc_body}
\end{equation}
where $\mathbf{C}_{n,t}^{b}$ is the coordinate transformation matrix from navigation to body frame at time $t$ and $\mathbf{p}^b$ is the position of the IMU in the body frame.

$\mathbf{v}_{t}^{b}$ can be computed given the velocity in the navigation frame $\mathbf{v}_{t}^{n}$ as
\begin{equation}
\mathbf{v}_{t}^{b} = \mathbf{C}_{n,t}^{b} (\mathbf{v}_{t}^{n} - \mathbf{v}_{c,v,t}^{n}) - \boldsymbol{\omega}^b_t \times \mathbf{p}^b
\label{eq:velocity_body}
\end{equation}
where $\mathbf{v}_{c,v,t}^{n}$ is the water current velocity surrounding the vehicle at time $t$.

Equation~\eqref{eq:damping} shows how the damping is defined given the linear and angular velocities at time $t$.
\begin{equation}
\mathbf{D}(\mathbf{\nu}_{t}, t) = \mathbf{D}_{l,t} \cdot \mathbf{\nu}_{t} + |\mathbf{\nu}_{t}|^{T} \cdot \mathbf{D}_{q,t} \cdot \mathbf{\nu}_{t}
\label{eq:damping}
\end{equation}
The linear damping matrix $\mathbf{D}_{l,t}$, the quadratic damping matrix $\mathbf{D}_{q,t}$ and the inertia matrix $\mathbf{M}_{t}$ are time dependent, since for all of them a sub matrix is part of the filter state.
The filter states $\mathbf{D}_{l,\text{sub},t}$, $\mathbf{D}_{q,\text{sub},t}$, $\mathbf{M}_{\text{sub},t}$ $\in \mathbb{R}^{2\times3}$ are defined by removing the rows 3 to 6 and columns 3 to 5 from the full damping and inertia matrices $\mathbf{D}_{l}$, $\mathbf{D}_{q}$, $\mathbf{M}$ $\in \mathbb{R}^{6\times6}$. In other words we model the $x,xy,x\psi,yx,y,y\psi$ terms of the matrices in the filter, where $\psi$ is the yaw. Because we expect them to have the major impact with respect to the horizontal accelerations and velocities, in case of an AUV keeping roll and pitch stable. It would be easy to extended the filter states and add more model terms, however it is a trade-off between the additional benefit, the computational complexity and potential filtering instability. 

The damping and inertia state prediction models have a base time varying component, with a timescale of around one hour, modeled as in \eqref{bias_equation}.
The vehicle parameters are initialized using a prior system identification, with the means of the states set at these values in the first order Markov process equation. Since the vehicle parameters are states in the filter, the systematic uncertainty in their error can be accounted for, which acts like a bias rather than a noise. This allows even a low accuracy system identification, or very crude estimates of the parameter values, to still allow estimation without resulting in overconfidence, due to the bias error having a stronger and different effect to simply increasing the uncertainty in the vehicle model noise. 
This also allows the vehicle modeling to adapt to different scenarios, such as surfacing or changes to the vehicle following the system identification, while constraining their value range by utilizing the first order Markov process model. 
Although these parameters could have been modeled as a constant, by allowing the parameters to have a time-varying component it acts as a way to implement ``model uncertainty''.
In this way, the robustness of the filter improves as we no longer fully trust our model to be a perfect representation of the true dynamics, which is most definitely the case with applying a simplified and computationally tractable model for real-time usage to the real-world.

\subsection{ADCP-aiding update equations}

Given the 3D velocities output from the ADCP, the observation function for each ADCP measurement is
\begin{equation}
\textbf{z}_{c,i}(t) = \textbf{C}_{n,t}^{b}(- \textbf{v}_{t}^{n} + \frac{d_{max}-d_{i}}{d_{max}}\textbf{v}_{c,v,t}^{n} + \frac{d_{i}}{d_{max}}\textbf{v}_{c,b,t}^{n}) + \textbf{b}_{a,t} + \nu_{a}
\label{z_adcp}
\end{equation}


where $\textbf{z}_{c,i}$ is the ADCP measured current vector in the i$^{th}$ measurement cell, $\textbf{C}_{n,t}^{b}$ is the coordinate transform from navigation/world frame to ADCP/body frame at time $t$,
$\textbf{v}_{t}^{n}$ is the vehicle velocity in the world/navigation frame, $\textbf{v}_{c,v,t}^{n}$ is the water current velocity surrounding the vehicle, $\textbf{v}_{c,b,t}^{n}$ is the water current velocity at the maximum ADCP range, $\textbf{b}_{a,t}$ is the bias in the ADCP measurement and $\nu_{a}$ is the random noise in the ADCP measurement, with a standard deviation given by the sensor manufacturer.

To reduce the state number of the filter, the vertical velocity of the water currents are not estimated. The ADCP measurement model is a depth dependent function with two water current states, which linearly interpolates between them. The states are located at the vehicle position, and at a water volume at end of the ADCP measurement range. The water velocity and the ADCP bias state prediction models have a base time varying component, with a timescale of around one hour for the water current and half an hour for the bias, modeled as in \eqref{bias_equation}. In addition to this, the water velocity state will vary more given spatial motion through a water current vector field. This component scales the process model uncertainty of the water velocity state according to the vehicle velocity. In this way, if the vehicle is slowly traveling through the water current vector field, it can account for the spatial scale of the water currents, which can depend on the environment. For example, water currents near complex bathymetry or strong wind and tides can contribute to smaller spatial scale water current velocity changes, compared to the case of the mid-water ocean \cite{medagoda2015autonomous}.

\section{Results}

%
%
%
%
%
%
%
%



All the experiments have been made using the \textit{FlatFish} AUV \cite{albiez2015flatfish} shown in Fig. \ref{fig:flatfish}.
As relevant sensors for our experiments, the vehicle is equipped with a KVH 1750 IMU, a Rowe SeaProfiler DualFrequency 300/1200 kHz ADCP/DVL, a Paroscientific 8CDP700-I pressure sensor, a u-blox PAM-7Q GPS receiver and six 60N Enitech ring thrusters. For heading evaluation purposes we also use a Tritech Gemini 720i Multibeam Imaging Sonar attached to the AUV.
The data sets have been collected during the sea trails of the second phase of the \textit{FlatFish} project close to the shore of Salvador (Brazil) during April 2017.

Since the experiments took place in the open ocean in all data sets, a fiber optic tether was attached to the vehicle for safety reasons. As a result, even though the vehicle model parameters were estimated with a prior system identification, there would be a large error associated with the model given the tether, so there is $\sim$20\% uncertainty in the parameter values. Nonetheless, the filter is robustly capable of accounting for this increase in the uncertainty of the vehicle model parameters. This allows the filter to adaptively change the parameters while keeping them in a constrained range through the use of the first order Markov process model.
The filter is capable of running in 14$\times$ real-time with an integration frequency of 100 Hz on computing available on the \textit{FlatFish} AUV.

\subsection{Heading estimation experiment}
\label{sec:heading_experiment}


In this data set we show that the filter is able to find its true heading without a global positioning reference, given an initial guess.
The mission consists of a initialization phase on the surface followed by a submerged phase before resurfacing. During the initialization phase the vehicle moves for around 8 minutes on a straight line in order to estimate its true heading and position by incorporating GPS measurements. In the submerged phase the vehicle changes its heading to face the target coordinate and follows a straight line for about 112 meters to reach it.

\begin{figure}[!h]
      \centering
      \includegraphics [width=0.45\textwidth,trim=0 0 0 5,clip] {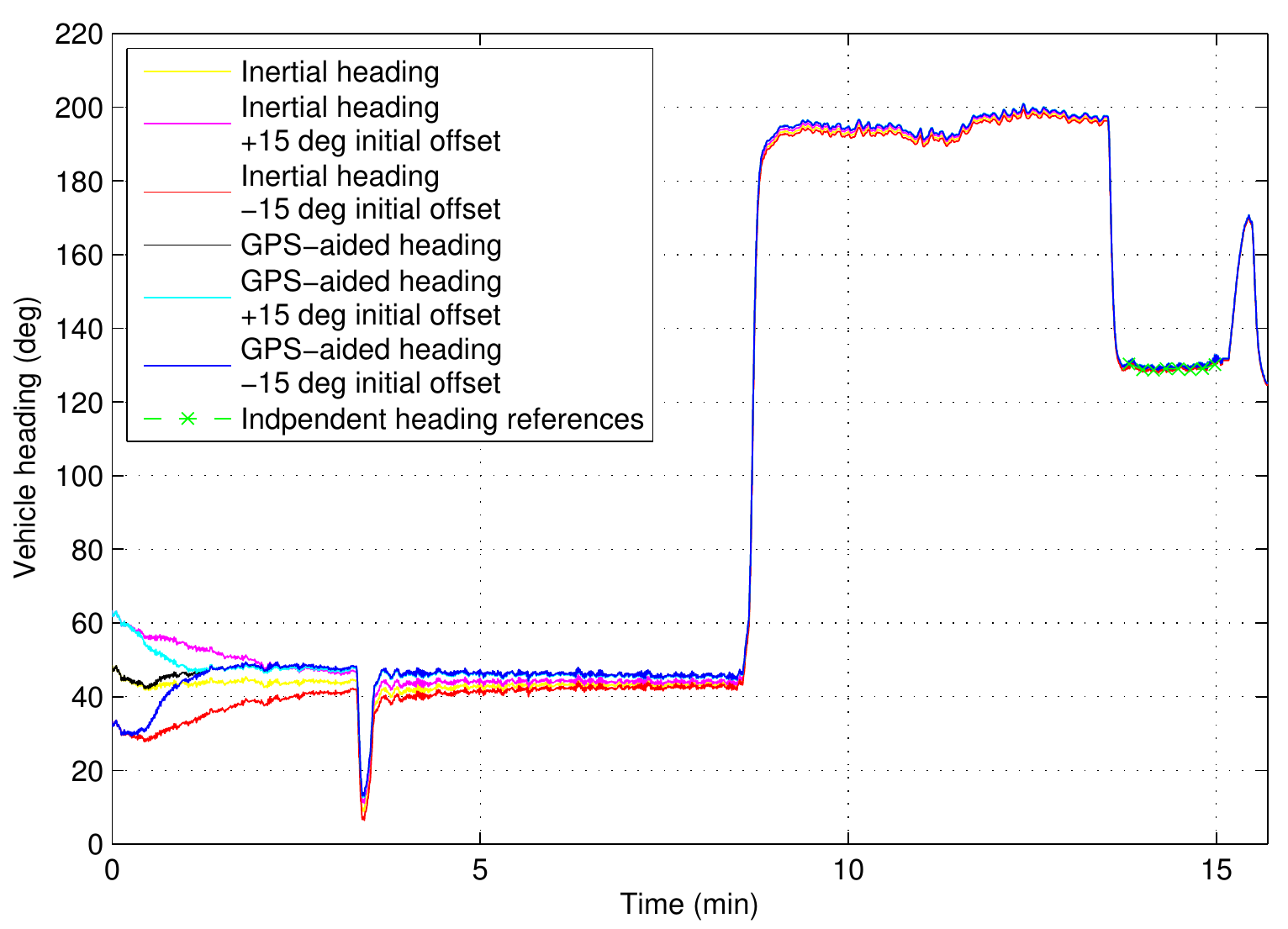}
      \caption{The plots show the estimated heading during the mission given different filter configurations and initial headings distributed over 30$^{\circ}$.
	       The green crosses show independent land mark based heading measurements. 200 seconds in the mission the heading offset was corrected resulting in the short change of attitude.}
      \label{fig:heading_comp}
\end{figure}

Fig. \ref{fig:heading_comp} shows six runs of the same data set in different filter configurations. Three GPS-aided runs with a different initial heading distributed over 30$^{\circ}$, one with a close initial guess (black line), one with a 15$^{\circ}$ positive offset (cyan line) and one with a 15$^{\circ}$ negative offset (blue line). Due to the help of the GPS measurements the estimated headings converge in the first 5 minutes.
The three runs not integrating a global position reference starting with the same heading distribution show that the filter is able to find its true heading by observing the rotation of the earth (gyro compassing), relying only on Inertial and velocities.
After 15 minutes the GPS-aided and the non-GPS-aided estimated headings have converged with an uncertainty below 0.5$^{\circ}$ (1$\sigma$).
Initial errors $>$15$^{\circ}$ will converge as well given more time. Critical however are initial errors close to 180$^{\circ}$.
The green crosses show multiple independent measurements of the expected vehicle heading based on landmarks (poles) visible in the multibeam imaging sonar on the vehicle.
The average difference between the landmark based headings to the filter estimates is below 1$^{\circ}$.
We expect the uncertainties of these measurements to be within 5$^{\circ}$ due to the uncertainties associated with the pole positions in surveyed maps and in the sonar images.

\begin{figure}[!h]
      \centering
      \includegraphics [width=0.45\textwidth,trim=0 0 0 5,clip] {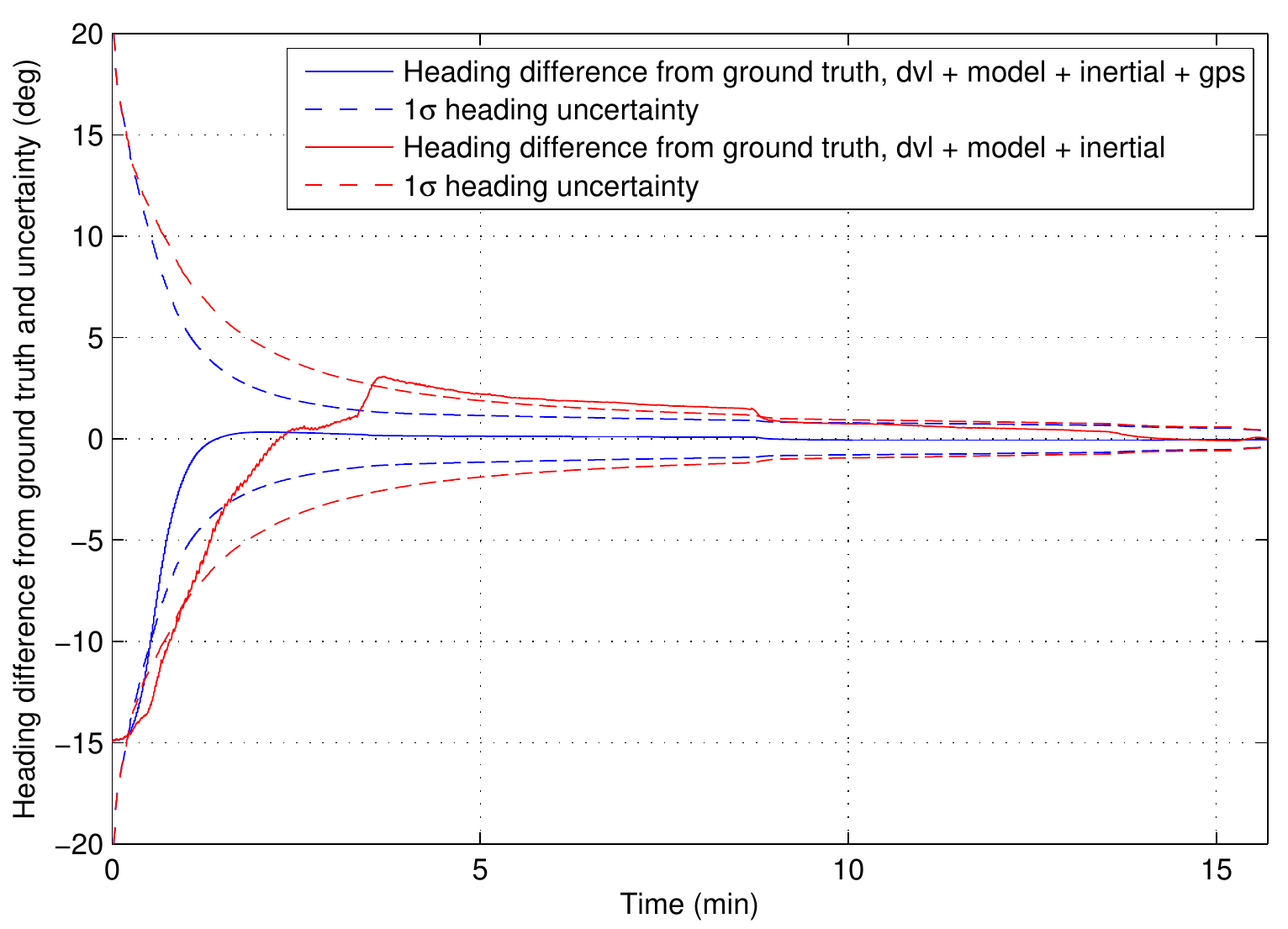}
      \caption{The blue solid line is the error in heading with integrated GPS measurements.
	       The red solid line is the error in heading without the integration of GPS measurements.
	       The dashed lines are the corresponding uncertainties (1$\sigma$).}
      \label{fig:heading_diff}
\end{figure}

Using the GPS-aided heading with a close initial guess shown in Fig. \ref{fig:heading_comp} (black solid line) as ground truth, we can have a closer look in Fig. \ref{fig:heading_diff} on the uncertainties and how the estimates improve.
In Fig. \ref{fig:heading_diff} both filter configurations start with an offset of -15$^{\circ}$ to the ground truth and an initial uncertainty of 30$^{\circ}$ (1$\sigma$).
The GPS-aided heading estimate converges, as expected, quickly to the ground truth while staying in the 1$\sigma$ bound.
For the heading estimate without global positioning reference we can see that the strong offset and high uncertainty in the beginning leads to a fast compensation in the correct direction with an overshot slightly exceeding the 1$\sigma$ bound. As the experiment progresses we can see that observing different orientations helps to estimate the gyroscope bias and therefore helps to detect the error between the expected rotation of the earth given the current orientation. We have shown that our filter is able to estimate its true heading by observing the rotation of the earth and that observations from different attitudes help to improve the process.

\subsection{Repeated square path experiment}


In this experiment we show how the filter performs when the vehicle travels a longer distance of 1 km without horizontal position aiding measurements, such as GPS.
The vehicle was following a 5 times repeated square trajectory with an edge length of 50 meter for $\sim$1 hour. After resurfacing, the position difference to the GPS ground truth is within 0.5\% of the traveled distance.

\begin{figure}[!ht]
      \centering
      \includegraphics [width=0.45\textwidth,trim=0 0 0 5,clip] {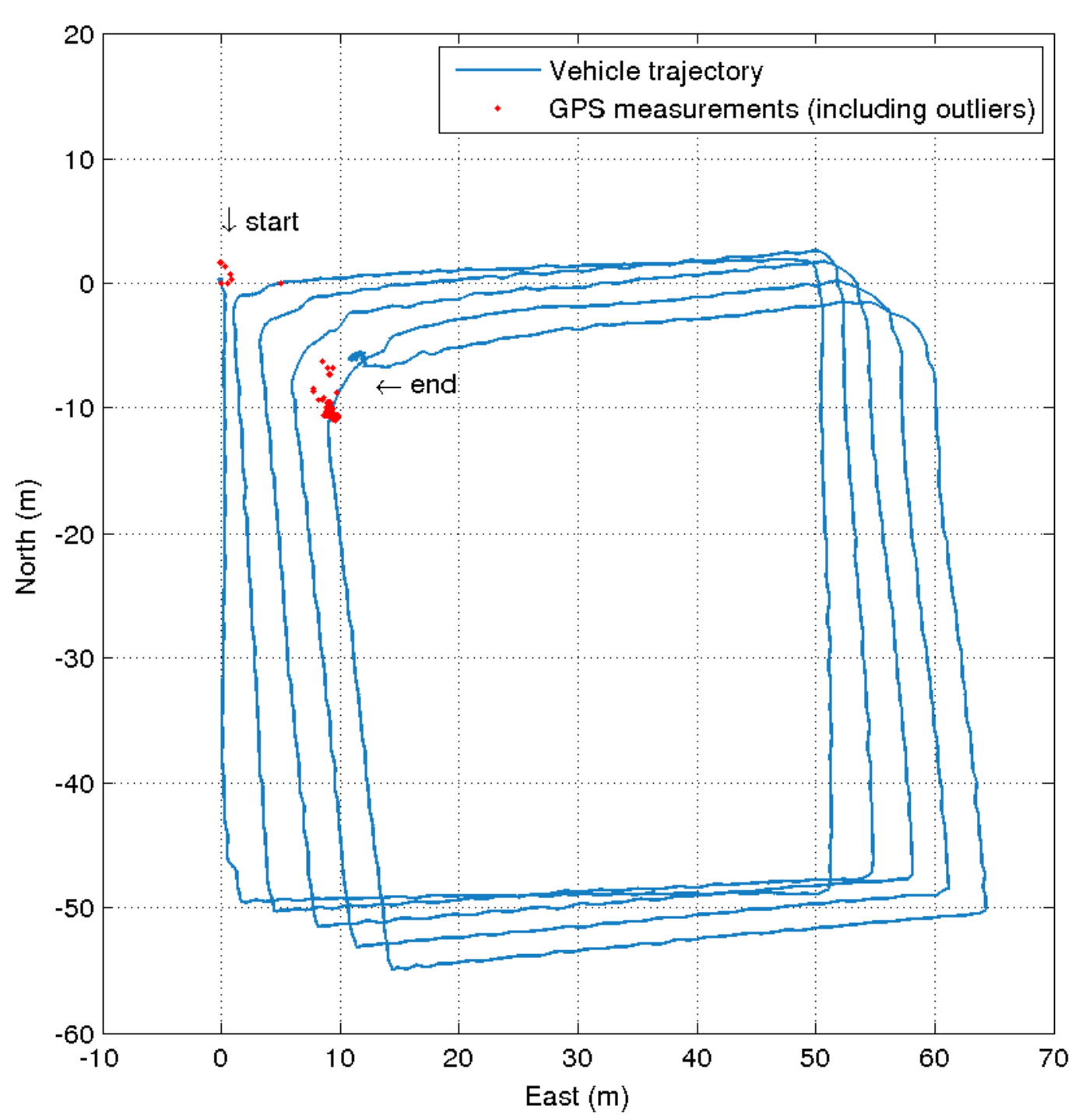}
      \caption{The blue solid line shows the trajectory of the vehicle performing 5 times a 50 meter square trajectory in a depth of 10 meter.
	       After traveling the distance of 1 km the horizontal (North/East plane) position difference is withing 5 meter (0.5\% of distance traveled).}
      \label{fig:square_trajectory}
\end{figure}

Starting with an initialization phase (same as in \ref{sec:heading_experiment}) on the surface, to estimate its heading and position using GPS measurements, the vehicle submerges to 10 m depth, performs the mission and surfaces at the end.
The blue line in Fig. \ref{fig:square_trajectory} shows the trajectory of the vehicle from minute 20 to minute 80 in the mission, i.e. 1 minute before submerging and 2 minutes after surfacing.
The red dots are the GPS measurements including outliers.

The pose filter used on the vehicle at the time the data set was created was not aware of the drift and the initial error in heading.
Our filter can correct the heading by observing the rotation of the earth and compensate for DVL dropouts utilizing the motion model.
However during the mission a fiber optic tether was attached to the vehicle which represents an unmodeled source of error.

\begin{figure}[!ht]
      \centering
      \includegraphics [width=0.45\textwidth,trim=0 0 0 -2,clip] {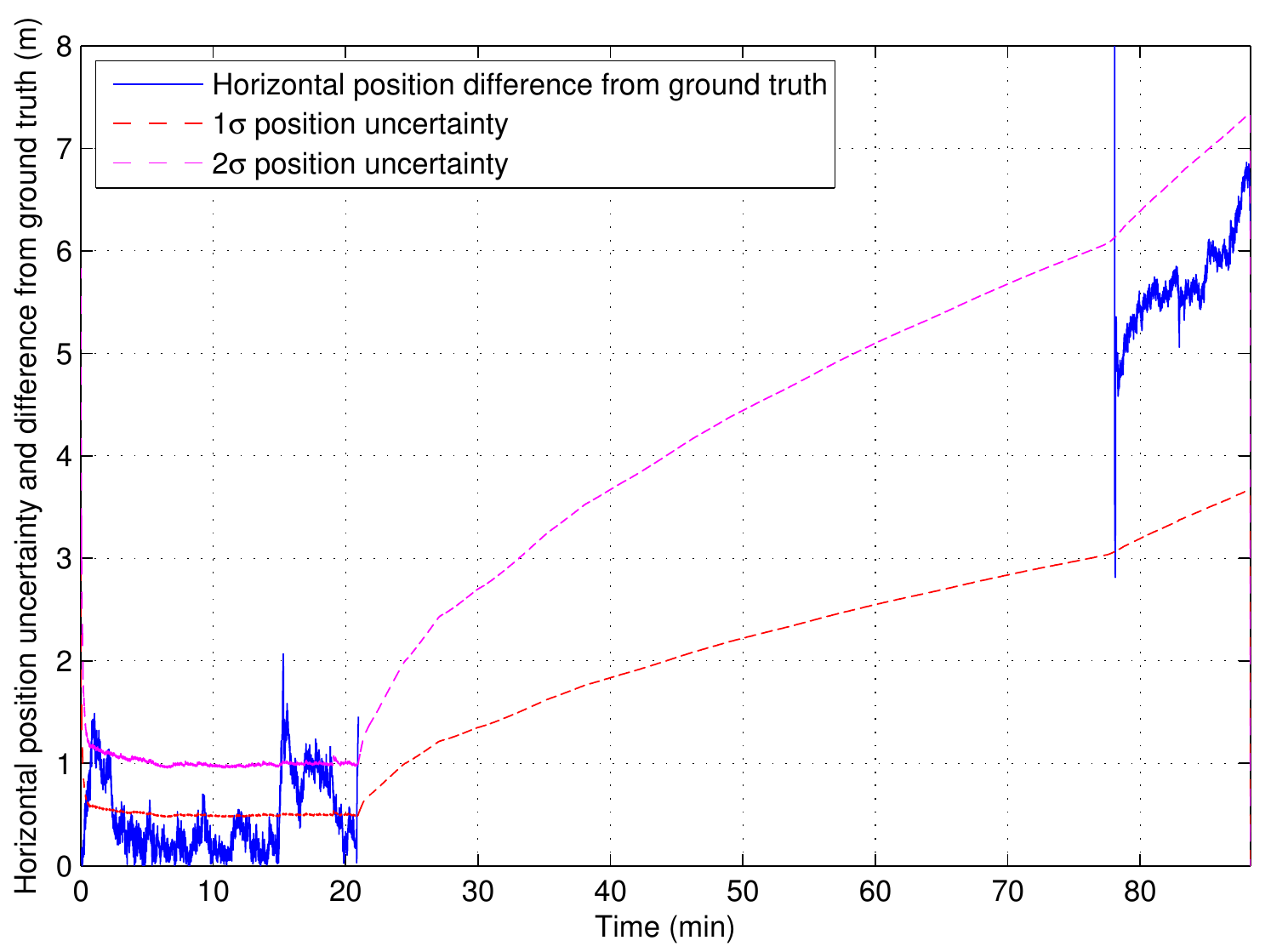}
      \caption{The blue solid line shows the horizontal (North/East plane) position difference with respect to the GPS measurements (including outliers).
	       The red and magenta dashed lines represent the corresponding uncertainty (1$\sigma$ and 2$\sigma$).}
      \label{fig:square_pos_diff}
\end{figure}

The blue line in Fig. \ref{fig:square_pos_diff} shows the position difference on the North/East plane with respect to the GPS measurements (including outliers).
During the first 20 minutes of the mission the GPS measurements are integrated in the filter allowing initialization.
After resurfacing (minute 78 and onward) the GPS measurements are not integrated allowing us to observe the difference to the ground truth.
After traveling a distance of 1 km the position difference is withing 5 meter (0.5\% of distance traveled) and in the 2$\sigma$ bound of the position uncertainty.

\begin{figure}[!ht]
      \centering
      \includegraphics [width=0.45\textwidth,trim=0 0 0 5,clip] {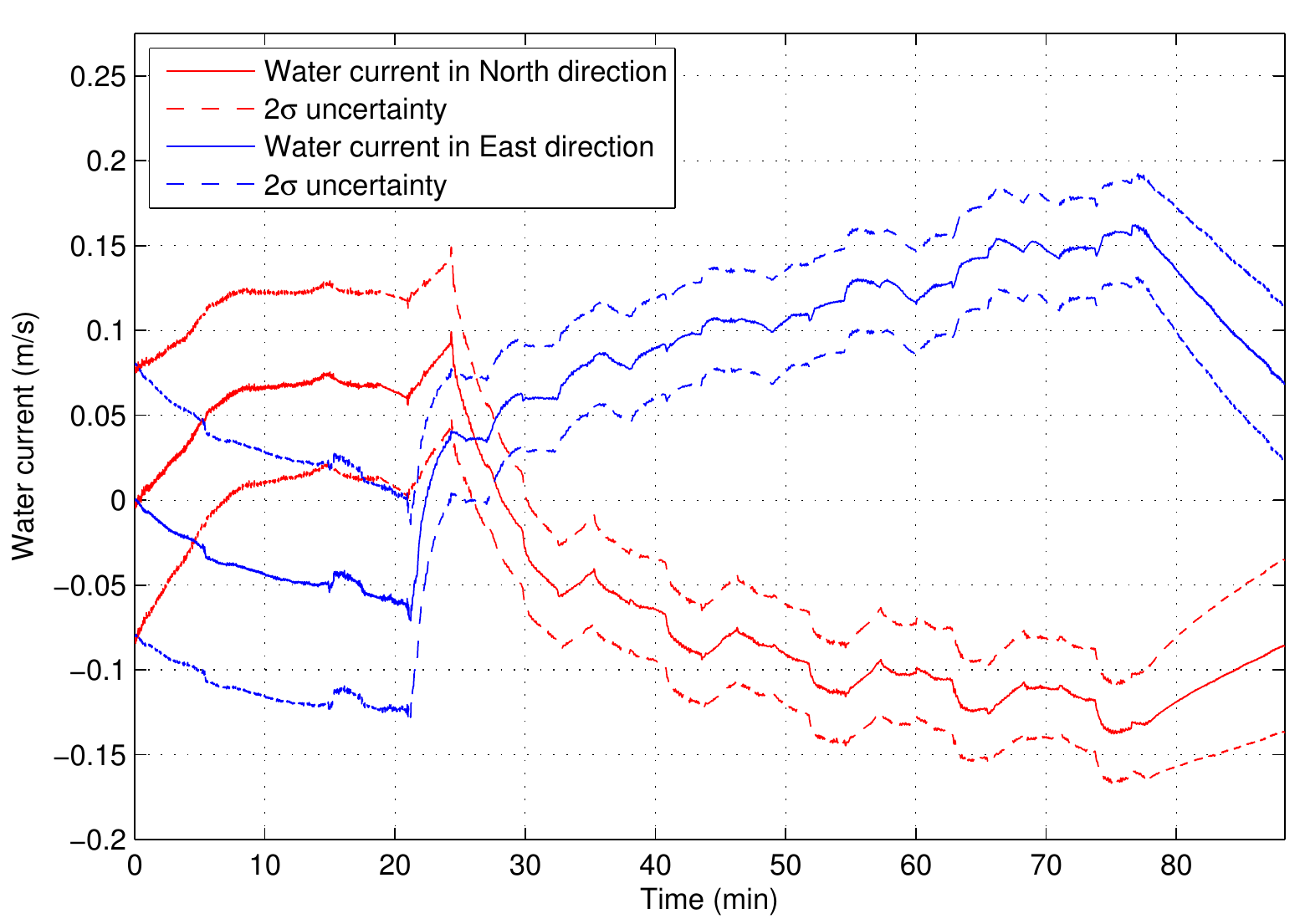}
      \caption{Estimated water current in north (red) and east (blue) direction. The dashed lines represent the corresponding uncertainties (2$\sigma$).}
      \label{fig:5x50m_square_water_current}
\end{figure}

In the case that ADCP measurements are not available the filter will estimate the water currents only by the difference between the motion model based velocity and the DVL based velocity, as modeled in \eqref{eq:velocity_body}.
Fig. \ref{fig:5x50m_square_water_current} shows the estimated water current velocities in North and East direction during this experiment without the aiding of ADCP measurements.
During the first 20 minutes the uncertainties of the water current velocities stay constant, since we apply the model-aiding measurements with an increased uncertainty in case the vehicle is surfaced.
When the mission starts and the vehicle submerges (starting around minute 21) to a depth of 10 meters we can see that the estimated water flow changes to the one on the surface and that its velocity continuously increases during the 1 hour mission. The uncertainties reduce during this phase since we trust the model more when submerged.
The impact of the tether attached to the vehicle is seen as an unmodeled but estimated drag, which changes depending on the direction the vehicle travels.

\begin{figure}[t]
      \centering
      \includegraphics [width=0.45\textwidth] {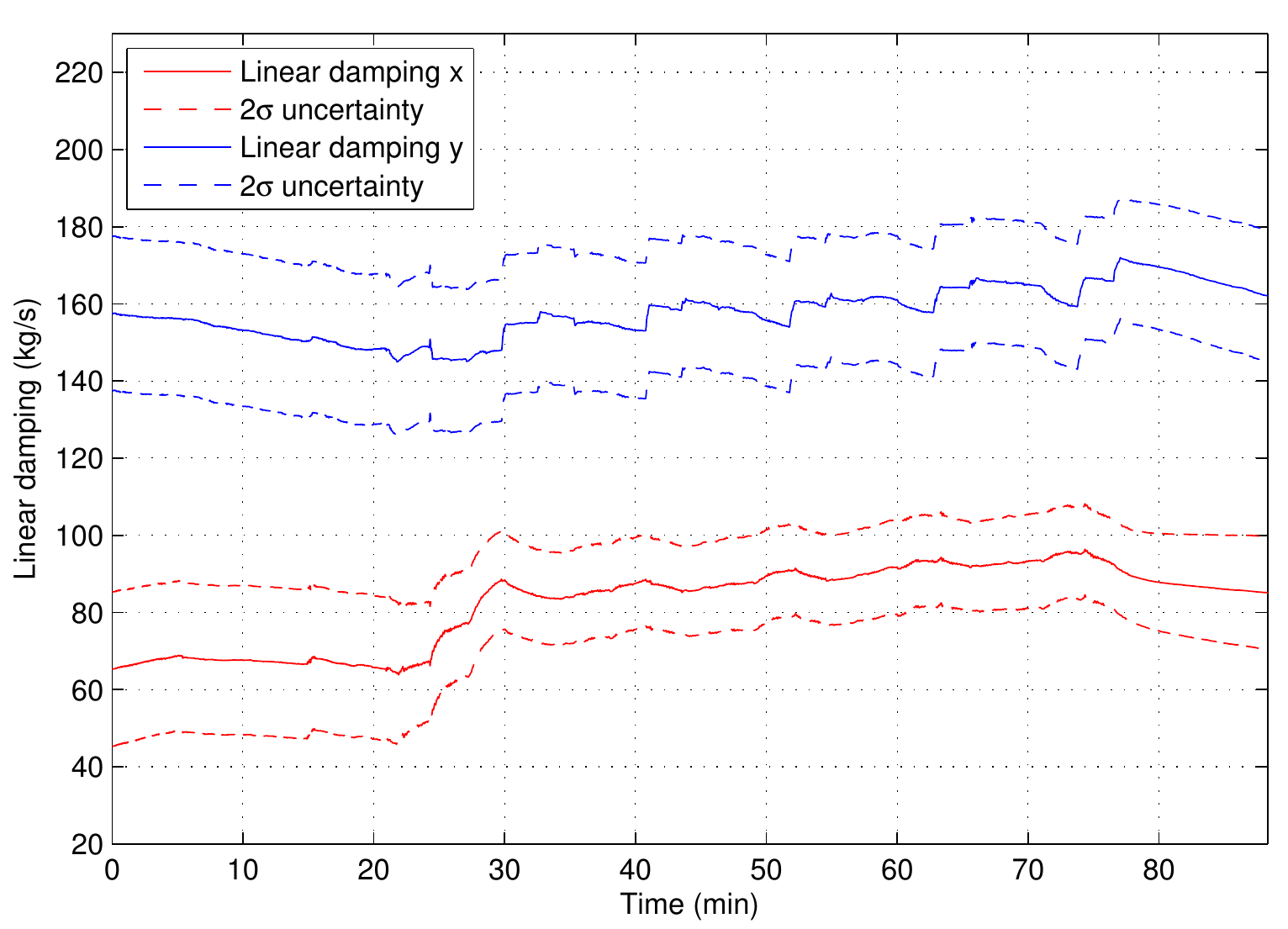}
      \caption{Linear damping in x (red) and y (blue) direction in the body frame. The dashed lines represent the corresponding uncertainties (2$\sigma$).}
      \label{fig:5x50m_damping}
\end{figure}

Fig. \ref{fig:5x50m_damping} shows the linear damping terms on the x and y-axis in the body frame of the vehicle and how they are refined during the mission.
Because the vehicle travels during the mission mainly in the forward direction, the damping term on the x-axis is refined and the corresponding uncertainty reduces more compared to the y-axis damping term. The uncertainty reduction reaches a limit however due to observability, and the first order Markov process model ensures that the parameters become neither overconfident nor unconstrained. In this way, the model parameters can adapt with time to new conditions and implicitly represents some uncertainty in the model equations themselves.


\subsection{Square path with ADCP}

\begin{figure}[!ht]
      \centering
      \includegraphics [width=0.45\textwidth,trim=5 10 5 25,clip] {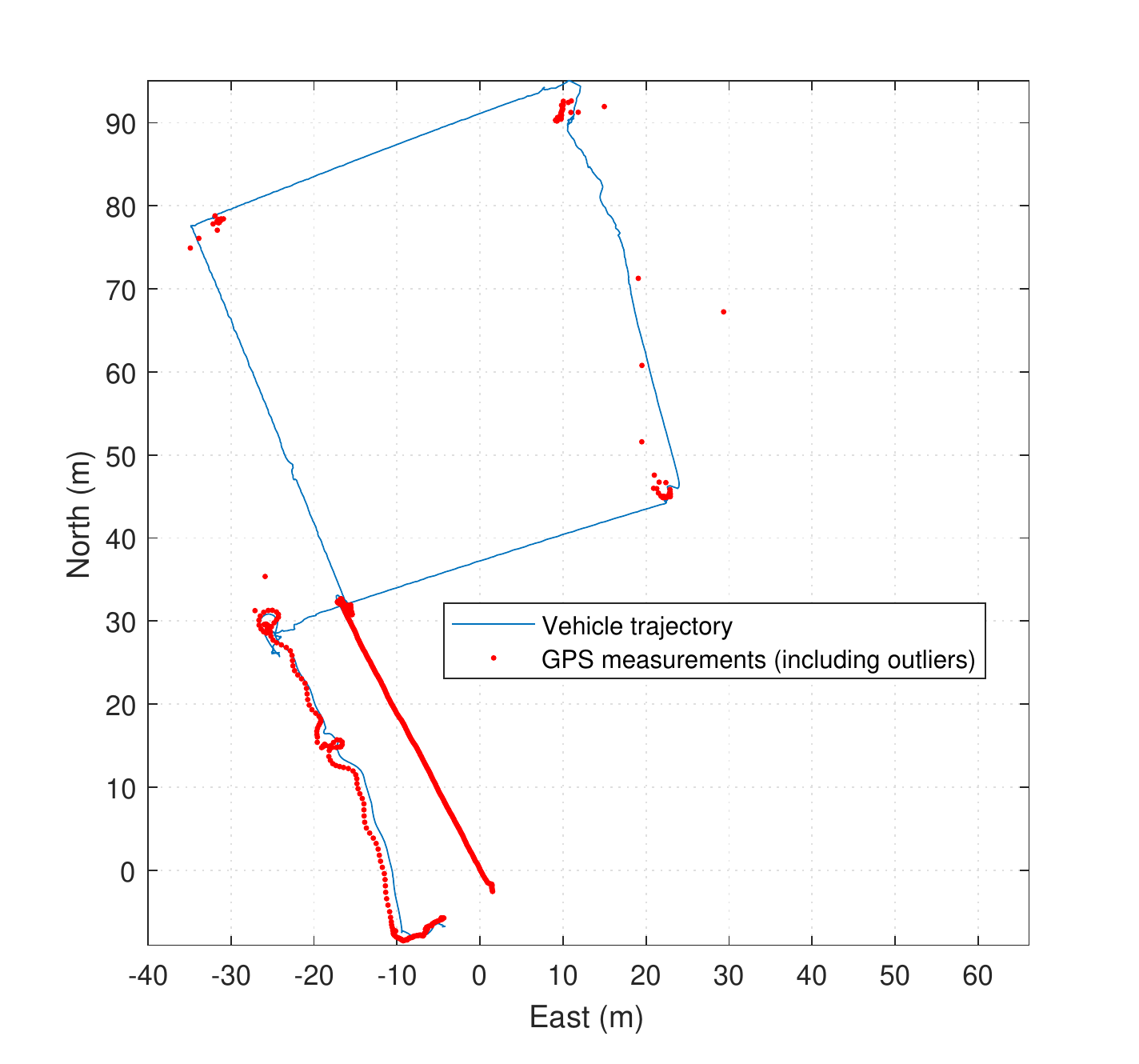}
      \caption{The solid blue line shows the trajectory of the vehicle performing a square path in a depth of 2 meter while surfacing in each corner.}
      \label{fig:adcp_traj}
\end{figure}

\begin{figure*}[!ht]
      \centering
      \includegraphics [width=0.75\textwidth,trim=0 5 0 0,clip] {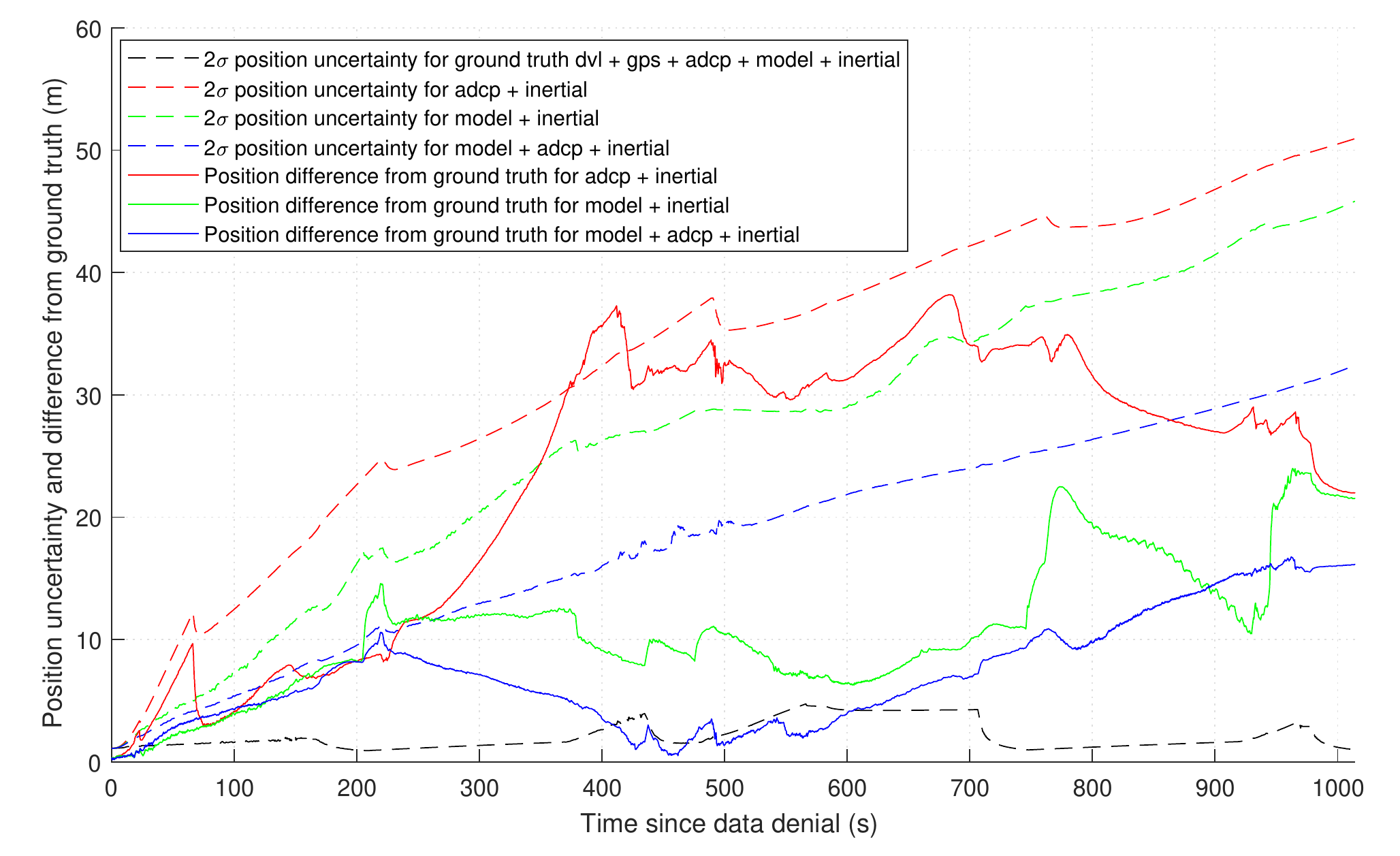}
      \caption{Square path with ADCP - The position uncertainties and differences from the ground truth are compared for different data denials.}
      \label{fig:adcp_compare}
\end{figure*}

\begin{table}[]
\caption{Filter position difference from ground truth and estimated uncertainty}
\label{adcp_table}
\centering
\begin{tabular}{|l|l|l|}
\hline
Filter measurements used & \thead{Estimated uncertainty\\after 1000 seconds} & \thead{Position difference \\from ground truth\\ after 1000 seconds} \\ \hline
Inertial + ADCP  & 50.9 m (2$\sigma$) & 22.0 m \\ \hline
Inertial + model & 45.7 m (2$\sigma$) & 21.5 m \\ \hline
Inertial + model + ADCP & 32.3 m (2$\sigma$) & 16.1 m \\ \hline
\end{tabular}
\end{table}

This mission undergoes a 600 second initialization phase on the surface (as in \ref{sec:heading_experiment}), then 1000 seconds of data denial to show the performance of the filter in different scenarios. During the data denial phase, the vehicle completes a square trajectory, and surfaces at the corners. The ground truth trajectory is shown in Fig. \ref{fig:adcp_traj}. The ground truth is determined using Inertial, DVL, GPS, ADCP and model-aiding.

Since this mission also includes ADCP measurements interleaved with DVL, the ADCP-aiding update is applied. During this mission, there are cases where the downward facing DVL drops out due to very low altitude (between 0-2m during the mission), and there is collision with the sandy bottom. Despite this challenging data set, the filter is capable of estimating the position of the vehicle, validated by the smooth trajectory without sudden corrections at the GPS measurements during the corner surfacing shown in Fig. \ref{fig:adcp_traj}.

With the full measurement filter (without data denial), the filter is able to handle DVL drop outs, which could be the case in low-altitude scenarios such as inspection or docking, by letting the model-aiding fill in during these time periods. Data denial further validates the filter performance in DVL loss scenarios, as shown in Fig. \ref{fig:adcp_compare}. In cases of DVL bottom-lock loss due to altitude being too high (simulated through data-denial), the ADCP and model-aiding combined gives the best solution, compared to either ADCP or model-aiding alone.

The position estimate differences compared to the ground truth for these data denials are consistent with the 2$\sigma$ uncertainty bounds, while remaining stable. At approximately 400 seconds following data denial, the filter with only ADCP and inertial measurements appears to slightly exceed the 2$\sigma$ bounds, due to a low altitude section with very little valid ADCP measurements, and some ADCP outliers are incorporated into the filter since the innovation gate increases due to inertial-only dead-reckoning. The ground truth also increases in uncertainty at this stage due to the lack of DVL measurements, relying more on the model-aiding. Following further measurements, the filter recovers and is able to reduce the difference between the filter estimate and the ground truth. This is possible since the water current estimate will not vary significantly in this timescale, so that the vehicle can use this state when there are ADCP measurements available again to estimate the velocity and thus position of the vehicle.

The ADCP-aiding typically performed worse in this case than the model-aiding, but this can be attributed to low altitude where there are very few valid ADCP measurements available. Nonetheless, incorporating these ADCP measurements into the model-aiding improved on the performance of either option. In addition to another source of velocity-aiding information from the ADCP, it also allows an independent source of information regarding the water currents surrounding the vehicle, which is required to transform the water relative velocity of the vehicle model to the navigation frame position used in the filter.

The results are further quantitatively compared in Table \ref{adcp_table}. The combination of the ADCP-aiding and model-aiding results in a significant improvement compared to model-aiding alone, reducing position uncertainty from 45.7m (2$\sigma$) to 32.3m (2$\sigma$) during 1000 seconds of data denial.

%


%
%

\section{Conclusions}


The filter designed and implemented in this paper would be appropriate for general AUV navigation, despite not using a navigation grade IMU. In comparison to \cite{hegrenaes2011model}, the primary insight to the design of this filter is the incorporation of the acceleration state, and adding many parameters as states to account for their correlated error, while modeling with a first order Markov process to constrain the change the filter can apply. The engineering design trade-off is that adding too many states will unnecessarily add computational complexity and potential filtering instability. 

This furthers the state-of-the-art for robust filter design for INS, model-aiding and ADCP measurements, capable of real-time performance, consistency and stability as outlined in the experiments, while remaining conceptually simple. 
This paper has shown a manifold based UKF that applies a novel strategy for inertial, model-aiding and ADCP measurement incorporation.
The filter is capable of observing and utilizing the Earth rotation for heading estimation to within 1$^{\circ}$ (2$\sigma$) by estimating the KVH 1750 IMU biases. 
The drag and thrust model-aiding accounts for the correlated nature of vehicle model parameter error by applying them as states in the filter. The usage of the model-aiding is validated through observing that the filter remains consistent and does not become overconfident or unstable in the real-world experiments, despite uncertain vehicle model parameters. 

It is hypothesized that the usage of time varying first order Markov processes to model these parameters act as a way to implement ``model uncertainty'', improving the robustness of the filter as we no longer fully trust our model to be a perfect representation of the true dynamics, which is most definitely the case with applying simplified and computationally tractable model for real-time usage to the real-world.
ADCP-aiding provides further information for the model-aiding in the case of DVL bottom-lock loss. The importance of water current estimation is highlighted in underwater navigation in the absence of external aiding, justifying the use of the model-aiding and ADCP sensor. Through data denial, scenarios with no DVL bottom lock are shown to be consistently estimated. Additionally this work was implemented using the MTK and ROCK framework in C++, and is capable of running in 14$\times$ real-time on computing available on the \textit{FlatFish} AUV.

Future work would include full spatiotemporal real-time ADCP based methods to more accurately model and observe the water current state around vehicle. This requires implementing a mapping approach, such as the work from \cite{medagoda2016mid} \cite{medagoda2015autonomous}. 
The primary source of bias uncertainty for the KVH 1750 IMU is due to temperature change. If the temperature of the IMU can be controlled, or this bias can be calibrated with further experiments, then the performance can be further improved. Further heading evaluation will be possible with better ground truth, such as a visual confirmation or by utilizing an independent heading estimator such as an iXblue PHINS, so that a more accurate heading comparison can be undertaken. The error in alignments of sensors could also be further compensated, perhaps by adding states to the filter similar to the strategy for other systematic biases. Finally, further experiments and implementations in a variety of scenarios are planned to further test and refine the proposed filtering strategy.





\addtolength{\textheight}{-12cm}   




\section*{Acknowledgement}

We like to thank Shell and SENAI CIMATEC for the opportunity to test the presented work on \textit{FlatFish}.

We also like to thank all colleagues of the \textit{FlatFish} team for their support and Javier Hidalgo-Carri\'o for his review.

This work was supported in part by the EurEx-SiLaNa project
(grant No. 50NA1704) which is funded by the German Federal
Ministry of Economics and Technology (BMWi).


\bibliography{Bib,jck}
\bibliographystyle{plain}

\end{document}